# A Study of the Minimum Safe Distance between Human Driven and Driverless Cars Using Safe Distance Model

Tesfaye Hailemariam Yimer[1], Chao Wen[1], Xiaozhuo Yu[2], Chaozhe Jiang[1]*

1  School of Transportation and Logistics, Southwest Jiaotong University, Chengdu, China
2  Amazon Toronto, Toronto, Canada

Email:   tsfhailemariam04@my.swjtu.edu.cn,   wenchao@swjtu.cn,   jamesonyu95@gmail.com,
          *jiangchaozhe@swjtu.edu.cn

## Abstract
When driving, it is vital to maintain the right following distance between the vehicles to avoid rear-end collisions. The minimum safe distance depends on many factors, however, in this study the safe distance between the human-driven vehicles and a fully autonomous vehicle at a sudden stop by an automatic emergency brake was studied based on the human driver ability to react in an accident, the vehicles' braking system performance, and the speed of vehicles'. For this approach, a safe distance car-following model was proposed to describe the safe distance between vehicles on a single lane dry road under conditions where both vehicles keep moving at a constant speed, and a lead autonomous vehicle suddenly stops by automatic emergency braking at an imminent incident. The proposed model then finally was being tested using MATLAB simulation and results showed that confirmed the effectiveness of this model and the influence of driving speed and inter-vehicle distance on the rear-end collision was also indicated as well compared with the two and three seconds rule of safe following distance.  The three – seconds safe distance following rules is safe to be applied for all speed limits; however, the two seconds can be used on speed limits up to 45 Km/hr. A noticeable increase in rear-end collision was observed according to the simulation results if a car follows a driverless vehicle with two seconds rule above 45 km/hr.

## 1. Introduction

Even though Road Traffic injuries rate is stabilized and declined relative to world population size and many vehicles, road traffic injuries are the eighth leading causes of deaths for people of all age globally and the leading cause of death for children and young adults 5 -29 years of age [1]. Nearly one-third (almost 33%) of all motor vehicle accidents are caused by the rear-end collision of automotive crashes [2]. Rear-end collisions are caused by errors and factors which described broadly as Human, Vehicle, and Roadway environments; however, human factors take a lion share for the occurrences of accidents [3]. An error made by human drivers may lead to the event, where a crash, near-crash or an incident to occur based on detection, decision, and the reaction of the action of the lead vehicle. Moreover, individual vehicle factors within traffic, which include speeds, braking capabilities, and driver behavior such as driving skill, perception of safety, and visual alertness, are also affecting road capacity and safety. Gradually, engineers developed different driving systems to reduce and eventually eliminate human errors causing crashes to happen [4].

As seen, the recently developed technologies, Autonomous vehicles that have the potential of transforming travel behavior, and the next generation of vehicles aimed at providing a cleaner and safer mode of transport and revolutionize the way we experience travel will soon share the existing road and road infrastructures. Besides, the current traffic rules and regulations will govern those fully autonomous vehicles along with human-driven vehicles. Having this in mind, every individual will think what impacts those autonomous vehicles, which have different driving behavior, may have brought on safety while interacting with human-driven vehicles. It might be thought that they will reduce the occurrences of rear-end collisions; however, it needs to be investigated to what extent and in what situations. Undeniably, it is, therefore, the research area to search answers for questions bore and/or will bear on every individual mind even lots of researches are currently going on.

The European New Car Assessment Program (Euro NCAP) performed standardizing tests on different autonomous vehicles with a constant speed of 20 – 60 km/hr relative to 5 km/hr and 8 km/hr pedestrian speed crossing from behind obstruction showed satisfactory results even if there is a difference among vehicles [5]. However, this test checked only the effectiveness of Automatic Emergency braking on avoiding forward collision. Here, the question is the safety distance about the follower vehicle not to collide with the rear bumper of the lead vehicle, rear-end collision if the follower is a human-driven vehicle. The Autonomous vehicles have the ADAS system which used relative position accuracy between the cars in a range of ±2 cm and the possibility of tests according to ISO15623 - forward vehicle collision warning systems and EURO NCAP - AEB procedure [6]helps to avoid forward-collision in return this performance will affect the follower vehicle.

For years, numerous studies have been conducted to propose different car-following models, which describe the operating safety of successive vehicles in moving [7]–[10]. Researchers have investigated the car following behaviors with different safe distance car following models considering different scenarios and situations and still there are researches on going dealing with the distance gap (spacing) of vehicles on moving and standstill which is of particular importance from the points of the safety of rear-end collisions through which drivers can react to take actions. However, all these models have been postulated for similar leader and follower vehicle types.

Therefore, the level of automation differences in car-following modeling with different scenarios has to be considered to analyze their impact on road traffic safety while it seems to be ignored in most of the previous models. The Human-driven (conventional vehicle) will collide with the rear end of the leading vehicle immediately ahead and argued to be responsible for these crashes unless the lead vehicles suddenly stopped with low driving speed for which it can react to any imminent incident in front.

In this study, since actual braking distance is affected by the type of vehicle's brake built-in, the inter-vehicle distance is investigated based on the relative speed of vehicles and reaction time in following and sudden stop states to understand the speed which causes for a rear-end collision.

This paper, therefore, concentrates only on modeling the safe distance between vehicles relative to their speeds and how reacting to the situations for reducing the possibility of human-driven vehicles making rear-end collision while a lead vehicle automatic emergency braking is in effect. It also contributes to check whether the existing speed limits in urban roads meet such conditions. Accordingly, it will also help to the revision of the vehicle following rules and speed limits.

The structure of the paper is organized as follows. Following the introduction, literature is reviewed under section 2, materials and methods for the car following procedure with safe distance model is described in section 3. The impacts of sudden automatic emergency braking of the leading vehicle on the following vehicle are simulated using MATLAB, and the results are explained and illustrated from the perspective of longitudinal clearance and speed in section 4. Finally, Results are summarized and concluded in the last section of the paper, in section 5.

## 2. Literature Review

Safe distance for follower vehicles is a great concern if vehicle automation is different, and when there is sudden braking by a front vehicle. The behavior of the following vehicle is affected by the type of the lead vehicle[11] which presented a car-following model that incorporates the effect of the lead vehicle type to predict the following vehicle's speed based on the relative speed and spacing and confirmed that the type of lead vehicle affects. However, the type of vehicle can have a significant influence on the following behavior, especially in heterogeneous traffic characterized by a mix of vehicles having different static and dynamic properties.

Brackstone *et al.* 2009 on their study confirmed that the driver-following behavior is affected by the type of the lead vehicle [12]. As Brackstone *et al.* 2009 mentioned, there are several reasons for the effect of a driver following behavior; a typical mix of heterogeneous traffic consists of vehicles such as cars, two-wheelers, three-wheeled auto-rickshaws, light commercial vehicles, buses, and trucks. These vehicles significantly vary in the static characteristics (such as length, width, and size) and dynamic characteristics (such as acceleration/deceleration and maximum speed). The automation differences of vehicles that brought variation in dynamic characteristics still have not yet been well addressed, especially the deceleration characteristics required to stop vehicles. The acceleration/deceleration characteristics will affect the gap needed for safe stopping and, thereby, the gap a vehicle maintains [13],[14].

As it has clearly been stated in most of the previous studies, on the mixed traffic, the following behavior of a driver depends on the types of leader and follower vehicles. There are only a few attempts made to study the effect of factors like; lead vehicle size, type, traffic composition, volume, headway distributions, and automation differences on longitudinal behavior of vehicles under mixed traffic conditions. In this study, the automation differences of vehicles considered as mixed traffic, for traffic safety are limited and crash safety, which impacts the following human-driven vehicles operating characteristics due to the interaction of fully autonomous vehicles on mixed traffic as lead vehicles are not still well-considered and analyzed. Hence, an attempt has been made to analyze and model the vehicle following behavior under mixed traffic conditions.

SAE International, the professional association of automotive engineers, identifies full automation as Level 5 while NHTSA as level 4, in which vehicles are capable of driverless operation in all circumstances, and both identify vehicles which have no autonomous features at all as Level 0[15],[16]. For this paper, level 5 and level 0 vehicles are used for modeling the new safe distance model as leading and follower vehicles, respectively, and used for the rest of the section of the paper.

For this approaches, we proposed a safe distance car-following model to describe the relationship between the speed of lead autonomous and human-driven follower vehicles to a minimum safe distance on a single lane dry road under conditions where a lead vehicle is moving at a constant speed with different driving situation and suddenly stopped by emergency braking at an imminent incident.

# 3. Materials and Methods
## 3.1. Safe Distance Model Notation

During a longitudinal drive, a safe distance is a critical maneuver to a follower car in a car-following model, which is determined by the relative speed, reaction time, and the maximum deceleration specific to the vehicle. Therefore, the Safe distance model in this paper study the inter-vehicle distance between the lead, autonomous vehicle, while the lead vehicle suddenly stops and moves with the constant speed, and Follower, a human-driven vehicle at or within a speed limit.

Figure 1 shows the notation of the safety distance, stopping, and critical safety distance between two successive vehicles at moving and a while sudden stopping state. We considered critical safety distance as a space left between vehicles at a complete stop. We should note that the safety distance varies with follower braking time based on its speed and maximum deceleration. When the follower driver perceived that the leader vehicle stops at a sudden, s/he starts to react and brake to stop a vehicle maintaining the safe distance or at least critical safety distance unless rear-end collision will happen. However, this is determined based on the space and time headways, relative speed of vehicles, and alertness of the driver.

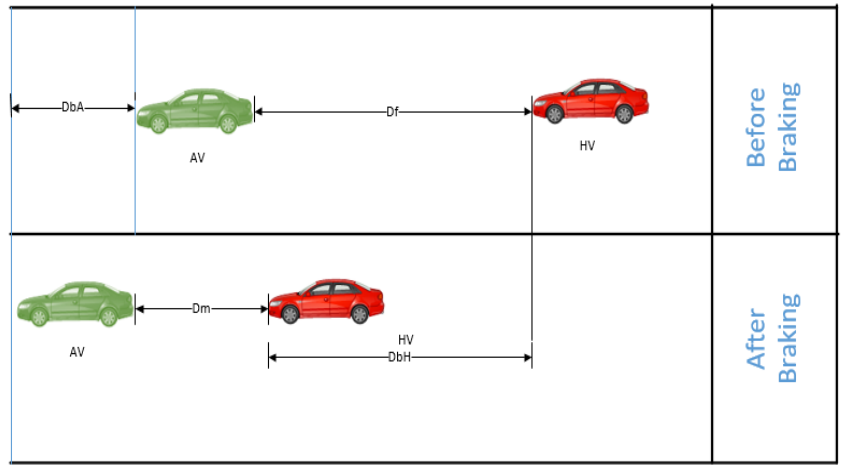

Figure 1. Safe distance model notation

In the model notation, the green car is a fully autonomous vehicle, which is a leading vehicle, and a Red car is a human-driven vehicle, which is a follower vehicle. $D_f$ is an inter-vehicle distance, the space between the rear bumper of the lead vehicle and front bumper of follower vehicle, $D_{bH}$ and $D_{bA}$ are the total stopping distances of human-driven and autonomous vehicles respectively and $D_m$ is the minimum or critical safety distance when both vehicles come to stop.

## 3.2. Safe Distance Modeling Description

A new safe distance model proposed based on the assumptions that always the leading vehicle is a fully autonomous and the follower is a Human-driven vehicle, entirely controlled by a human driver. Both vehicles are driven within or at the speed limit, and sudden braking of an autonomous vehicle with the help of automatic emergency braking (AEB) for preventing sudden bolting of road users from an accidental crash, is applied on a single lane dry road. We also assumed the autonomous vehicle displays the current speed and their autonomy level through the rear window screen to warn the follower car driver to be alert while following them.

A study of car-following model based on minimum safe distance was done by assuming that both leading and follower vehicles have the same braking process and considering the state of the leading vehicle [17] and have made an analysis comparing with the traditional safety models and safe distance models based on headway and braking process. However, in the proposed model, we used their assumptions with some modifications to study the interaction of human-driven and Autonomous vehicles considering the automation differences for braking performance.

Hence, in our model, we adapted their model to include such mentioned differences, which will work for all car following vehicles safe distance model driving at and within a speed limit. So our new safety distance model for the mixed vehicles of interactions having different braking performances adapted from their model [17] is described based on the above notations as:

$$D_f = (D_{bH} + D_m) - (D_{bA}) \qquad (1)$$

Where, $D_f$ is the inter-vehicle distance, safe distance, between the follower and leader vehicles, $D_{bH}$ and $D_{bA}$ are the total stopping distances of follower and lead vehicles respectively, $D_m, 2-5\ meters,$ 2 meters in our case, is the critical safety distance and L is the length of vehicles, 3.5 meters.

### 3.2.1. Braking Distance Using Braking Performance
### 3.1.1.1. Braking Distance of a Follower Vehicle

As mentioned in the model of Xu, L., et al. they used the changing deceleration value to calculate the braking distances of a car which is divided into three stages as reaction time stage when the vehicle is moved with constant speed, linearly increasing deceleration stage till maximum braking deceleration is attained and constant braking deceleration stage

**Reaction Stage**

A driver traveling at or near the design speed is more alert than one traveling at a lesser speed (AASHTO, 2001). From here, we assumed that the reaction time for the drivers of the Human-driven vehicle is dependent on how the alerted drivers react. However, depending on driver differences in an emergency, the most crucial factor for the vehicle stopping distance varied at a given vehicle speed in the range and used in this proposed model $tr_H = [tr_{Hmin}, tr_{Hmax}]$.

In this stage, the vehicle speed will not change, and therefore the car is still moving at an initial speed $V_H$. The driver reaction time is considered from the point s/he noticed the leader is breaking with an AEB. Therefore, $tr_H$ is considered as the total reaction time for following vehicle including the braking coordination time altogether, 0.8 to 1.5 seconds[18][19], and the reaction distance, $D_{trH}$ is:

$$[D_{trHmin,}\ D_{trHmax,}] = [V_H t_{rHmin}, V_H t_{rHmax}] \qquad (2)$$

**Linearly Increasing Deceleration Stage**

At the time of deceleration growth $t_l$, which is related to the performance of the follower car brake and its value is 0.1 - 0.2 seconds, 0.1 is taken for our case, braking deceleration is growing linearly from zero to achieve its maximum when the car brakes in the response of the leading vehicle action. We assumed that the relationship between the speed changes at any time with the following expression as:

$\frac{dv}{dt} = \beta t, where,\ \beta = -\frac{a}{t_l}$, then the speed of the follower car $v(t)$ at any time $t_l$ is:

$$v(t) = v_H + \frac{1}{2}\beta t^2 \qquad (3)$$

Substituting $\beta$ on the above equation, we obtain:

$$v(t) = v_H - a_{maxH}\frac{t^2}{2t_l} \qquad (4)$$

the distance a car moved at this stage is obtained by integrating the above equation becomes:

$$D_{t_l} = v_H t_l - a_{maxH}\frac{t_l^2}{6} \qquad (5)$$

where, $a_{max_H}$ is the maximum deceleration of the follower vehicle which is in the range 6 - 8 ms$^{-2}$ [20], [21], but we have taken, 7 ms$^{-2}$ and $V_H$ is the follower vehicle initial speed, which is at or within the speed limit.

**Constant Braking Deceleration Stage**

In this stage, the car moves at constant deceleration $a_{maxH}$ in the continuous braking time $t_{cd}$. Based on the kinematics, the braking distance is proportional to the square of the speed, and therefore the constant braking deceleration distance is:

$$D_{t_{cd}} = \frac{V_H^2}{2a_{maxH}} \quad (6)$$

Therefore, the sum of distance at each stage is the total braking distance a car moved:

$$D_{bH} = D_{trH} + D_{t_l} + D_{t_{cd}} \quad 7)$$

Then considering the driver reaction time and omitting the value of $a_{maxH}\frac{t_l^2}{6}$ in the linearly decelerating stage since its value is minimal, the total braking distance of the human-driven vehicle is re-written as:

$$[D_{bHmin}, D_{bHmax}] = [V_H(t_{rHmin} + t_l) + \frac{V_H^2}{2a_{maxH}}, V_H(t_{rHmax} + t_l) + \frac{V_H^2}{2a_{maxH}}] \quad (8)$$

### 3.1.1.2. Braking Distance of a Leading Vehicle

The green( fully autonomous or lead) car Central Processing Unit( CPU) senses a hazard ahead within 0.001 – 0.1 seconds and reacts accordingly within 0.01 – 0.1 seconds [22]–[24]. In our model, we assumed the minimum actuation time, which is 0.01seconds, and took the same time for environment sensing along with all sensor synchronization for decision. Summing up, all we took and assumed a reaction time as 0.02 seconds. Therefore, the braking distance for the leading car is:

$$D_{bA} = V_A t_{rA} \quad (9)$$

Where; $D_{bA}$ is the total stopping distance a green car moved during sudden braking, $V_A$ is the green car initial speed, which is a speed limit in this model; $tr_A$ is the minimum sum of reaction time for sensing the environment, perceiving to make a decision and actuation times to take control.

### 3.1.2. Modeling Safe Distances

Relying on the above braking performances of the cars, the safe distance model is going to be analyzed using different scenarios at uniform speed within and at the speed limit based on the state of the leading vehicle and braking processes at a sudden stop where the automatic emergency braking (AEB) is in effect.

The speed of the Leading car in this model is a posted speed limit since fully autonomous vehicles are capable of detecting the posted speed limit with their inbuilt technologies and moving accordingly. Even if the braking distance with AEB at an emergency stop for fully autonomous vehicles is minimal, we used the relative speed $V_{HA} = V_A - V_H$ for measuring the safe distance between the vehicles since our concern is Human Driven vehicle. Where $V_{HA}$ is a relative speed, $V_A$ is the initial speed of the leading car and $V_H$ is the initial speed of the Human-driven car. The reaction time of both vehicles also plays a significant role, and hence the relative reaction time $\Delta tr = tr_H - tr_A$ is considered Where $tr_A$ is the leading car reaction time which is vastly smaller than the Human-driven car driver's reaction time, $tr_H$.

In modeling the safe distance, we assumed that the leading vehicle with an emergency event indicator displays warnings to the follower vehicle when approaching the safe distance through the rear window screen and the speed of the leading car too. Besides, the detection range of Autonomous vehicles is considered for the speed change of human-driven cars before reaching the speed limit zone for all models except for the model where both are driven at an equal speed limit.

### 3.2.2.1 Safe Distance Modelling when both Vehicles are moving at a constant speed

In this case, both vehicles are only running at their initial speed unless there is a need for deceleration of the follower car and can safely follow with its initial speed. Therefore, the safe distance between the vehicles could be analyzed based on their initial speed variation only when the speed of the follower car is higher than the leader car's speed ($V_H > V_A$). All other cases could not be considered in our model since we thought they are safe to follow a car, which is driven at a constant speed and in a condition where there is no possibility to increase the speed of the follower car.

Rear-end collision will happen only if the follower vehicle moves at its initial speed at some time; therefore, the following vehicle should decelerate to maintain at least a minimum safe distance. Since in the model, we considered that the leading vehicle could not accelerate, only the follower vehicle has to decrease its speed to the leading vehicle. Therefore, the driving distance of the follower vehicle is:

$$D_{bH} = V_H(tr_H + t_l) + \frac{V_H^2 - V_A^2}{2a_{maxH}} \qquad (10)$$

While the following vehicle is braking to maintain the safe distance, the leading vehicle keeps moving at its initial speed. Then the distance the leading vehicle moved during this time is:

$$D_{bA} = V_A(tr_H + t_l + \frac{V_{HA}}{a_{maxH}}) \qquad (11)$$

Therefore, the minimum safe distance between the two vehicles is:

$$D_f = V_{HA}(tr_H + t_l) + \frac{V_{HA}^2}{2a_{maxH}} + D_m \text{, thus } D_f \text{ is in the range of } [D_{f_{min}}, D_{f_{max}}]:$$

$$df = \begin{bmatrix} V_{HA}(tr_{Hmin} + t_l) + \frac{V_{HA}^2}{2a_{maxH}} + D_m, \\ V_{HA}(tr_{Hmax} + t_l) + \frac{V_{HA}^2}{2a_{maxH}} + D_m \end{bmatrix} \qquad (12)$$

### 3.2.2.2 Safe Distance Modelling at an Emergency Stop of Leading Vehicle

In this case, the capability of leading vehicles stopped at a sudden with an AEB, following vehicle speed as well the driver's ability to react for a sudden state of change greatly affects the minimum safe distance between the two vehicles. Therefore, we assumed three different states to model the safe distance.

**Scenario- I: Speed of the follower vehicle is greater than the leader vehicle's speed ($V_H > V_A$)**

In this scenario, a collision may happen at a certain time if the follower and leader vehicles continue to move with their initial speeds, or it may happen if the lead vehicle stopped at an imminent incident while the follower vehicle keeps moving with its initial speed. Therefore based on the state of the leading vehicle and the braking performance of vehicles, the follower vehicle driver has to react to avoid a collision.

At a time of leading vehicle sudden braking, the leading vehicle moves a distance $V_A tr_A$ and the follower vehicle moved a distance, $V_H tr_A$. Following, the follower vehicle driver has to react and bring the vehicle to stop. Therefore, the follower vehicle moved a distance:

$$D_{bH} = V_H tr_A + V_H(tr_H + t_l) + \frac{V_H^2}{2a_{maxH}}$$

Then, the minimum safe distance between the two vehicles is:

$$D_f = V_{HA} t_{rA} + V_H(tr_H + t_l) + \frac{V_H^2}{2a_{maxH}} + D_m \qquad (13)$$

$D_f$ is determined by drivers' response to the state of leading vehicle condition. Therefore, whatever the driver is alert, there is variation among drivers and hence dependent on the reaction time in the range $[tr_{min}, \ tr_{max}]$. Thus, $D_f$ is in the range of $\left[D_{f_{min}}, D_{f_{max}}\right]$:

$$df = \begin{bmatrix} V_{HA}t_{rA} + V_H(tr_{Hmin} + t_l) + \frac{V_H^2}{2a_{max_H}} + D_m, \\ V_{HA}t_{rA} + V_H(tr_{Hmax} + t_l) + \frac{V_H^2}{2a_{max_H}} + D_m \end{bmatrix} \qquad (14)$$

**Scenario- II: Speed of the follower vehicle is equal to the leader vehicle's speed ($V_H = V_A$)**

When the follower speed is equal to the leader, the relative speed is becoming zero, which means $V_A$ and $V_H$ are the same. Therefore, the follower who alertly follows the leading vehicle can react to stop the vehicle following the leading vehicle action. Then, the inter-vehicle distance becomes:

$$D_f = V_H(tr_H - tr_A) + \frac{V_H^2}{2a_{max_H}} + D_m \qquad (15)$$

Since $D_f$ is therefore determined by the reaction time in the range, $[tr_{min}, tr_{max}]$

$D_f$ is in the range $\left[D_{f_{min}}, D_{f_{max}}\right]$ and it becomes:

$$D_f = \begin{bmatrix} V_H(tr_{Hmin} - tr_A) + \frac{V_H^2}{2a_{max_H}} + D_m, \\ V_H(tr_{Hmax} - tr_A) + \frac{V_H^2}{2a_{max_H}} + D_m \end{bmatrix} \qquad (16)$$

**Scenario- III: Speed of the follower vehicle is less than the leader vehicle's speed ($V_H < V_A$)**

This scenario is the safest case than others since we assumed that there is a bigger inter-vehicle distance between the two vehicles. However, if the follower driver keeps following the leading vehicle at its initial speed and the leader decelerates or stops, a collision may occur later on.

Assuming that the leading vehicle suddenly stopped with an emergency braking and moved a distance, $D_{bA} = V_A t_{rA}$. During the time, $t_{rA}$ when the leader vehicles come to a stop, the follower vehicle driver drives a car with an initial speed, and its distance becomes:

$$D_{bH} = V_H t_{rA} \qquad (17)$$

Since the leading vehicle comes to a stop within a fraction of seconds and we assumed its later state of condition at static state, the follower vehicle has to brake, and its braking distance is:

$$D_{bH} = V_H(t_{rH} + t_l) + \frac{V_H^2}{2a_{max_H}} \qquad (18)$$

Therefore, the total distance a follower car moved is:

$$D_{bH} = V_H t_{rA} + V_H(t_{rH} + t_l) + \frac{V_H^2}{2a_{max_H}} \qquad (19)$$

Then the safe distance between the vehicles is:

$$D_f = V_H(t_{rH} + t_l) - V_{HA}t_{rA} + \frac{V_H^2}{2a_{max_H}} + D_m \qquad (200)$$

Since $D_f$ is, therefore, determined by the reaction time in the range $[tr_{min}, \ tr_{max}]$, it becomes:

$$df = \begin{bmatrix} V_H(tr_{Hmin} + t_l) - V_{HA}t_{rA} + \frac{V_H^2}{2a_{max_H}} + D_m, \\ V_H(tr_{Hmax} + t_l) - V_{HA}t_{rA} + \frac{V_H^2}{2a_{max_H}} + D_m \end{bmatrix} \qquad (211)$$

## 4. Simulation Analysis Results and Discussions

We preferred to use both two and three seconds safe distance car-following rules within the speed limit on urban street single lane straight road for which no takeover is allowed, and the driverless car is always a leading vehicle for simulation analysis and discussion using MATLAB simulation Software.

For all simulation analysis, we used $[(t_{rhmin} + t_l), (t_{rhmax} + t_l)] = (0.9, 1.5 \text{ seconds})$. The two and three seconds' rule of safe following distance in table 1 is used for comparison with the model result for the speed range of 20 up to 60Km/hrs.

*Table 1: Seconds' rule safe following distance*

| Safe distance (m) | Follower Car Speed(Km/hr) | | | | | | | | |
|---|---|---|---|---|---|---|---|---|---|
| | 20 | 25 | 30 | 35 | 40 | 45 | 50 | 55 | 60 |
| Two-seconds | 11.11 | 13.89 | 16.67 | 19.44 | 22.22 | 25 | 27.78 | 30.56 | 33.33 |
| Three-second | 16.67 | 20.83 | 25.00 | 29.17 | 33.33 | 37.5 | 41.67 | 45.83 | 50.00 |

### 4.1. Safe Distance Modelling at a constant speed State

In this case, both vehicles are moving at their initial speed, and we considered the higher the follower vehicle speed. For all tested relative speeds, the leading vehicle was set initially at 20 km/hrs and invariable. The simulation was executed at 5 Km/hrs increment of relative speed using equation (12), and the results data are shown in table 2.

As shown in Table 2, in either of the low or high relative speed differences the time needed to stop a follower car is not sufficient enough other than the first instant of two and three seconds. It showed that rear-end collisions would happen at a certain time.

*Table 2: Safe Distance Simulation Data at uniform Speed state*

| Safe distance (m) | Relative Speed(Km/hr) | | | | | | | | |
|---|---|---|---|---|---|---|---|---|---|
| | 5 | 10 | 15 | 20 | 25 | 30 | 35 | 40 | 45 |
| d_f_min | 3.39 | 5.05 | 6.99 | 9.20 | 11.69 | 14.46 | 17.50 | 20.82 | 24.41 |
| d_f_max | 4.22 | 6.72 | 9.49 | 12.54 | 15.86 | 19.46 | 23.33 | 27.49 | 31.91 |

However, for the rest of the follower vehicle speeds, the distance is much more than the inter-vehicle distance of the model. Therefore, for a follower human-driven car, based on the simulation result, the rear-end collision is unavoidable if the speed is higher than the leader and continues moving at constant speeds.

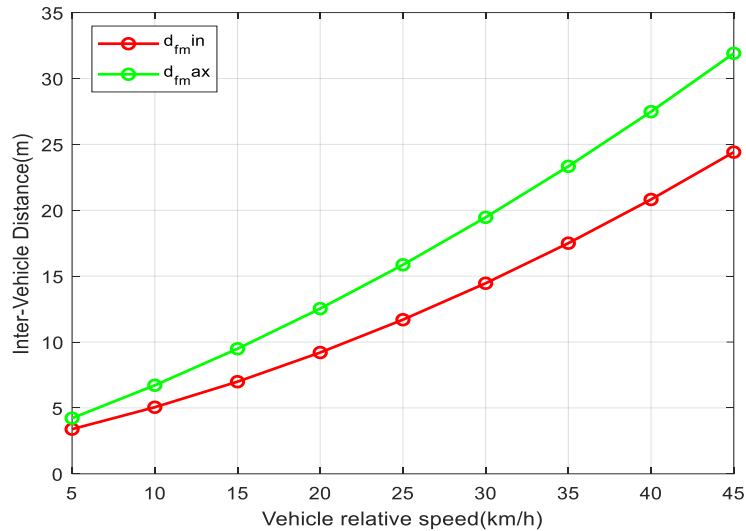

*Figure 2. Simulation of safe distance when HV speed is greater than AV at uniform Speed state*

As shown in figure 2, as the relative speed is increased the distance between vehicles has also increased. At lower relative speeds the driver needs less time to react and stop the vehicle before colliding with the in front vehicle, however for the higher relative speeds, the driver's reaction is dependent based on the spacing with the lead vehicle to adapt the follower car speed. The simulation result confirmed that larger spacing is needed for larger speed differences for which before a driver observes the difference. Since in a uniform motion, the relative speeds increment is also proportional to the safe distance, what matters is the speed of the follower vehicle and the state of conditions of the lead vehicle. Therefore, the following vehicle speed can be kept constant until the driver observing the speed differences and has to react accordingly to maintain a safe distance.

### 4.2. Safe Distance Modelling at a Sudden Braking of Leading Vehicle

#### 4.2.1. Speed of the follower vehicle is greater than the leader vehicle's speed ($V_H > V_A$)

For the effect of follower speed towards a collision, the leading vehicle was set initially at 20 km/hrs and invariable. The simulation was executed at 5 Km/hrs increment of follower vehicle speed using equation (14), and the results data are shown in table 3, and figure 3 and analysis are made in comparison with the two and three-second rule of safe following distances(Table 1).

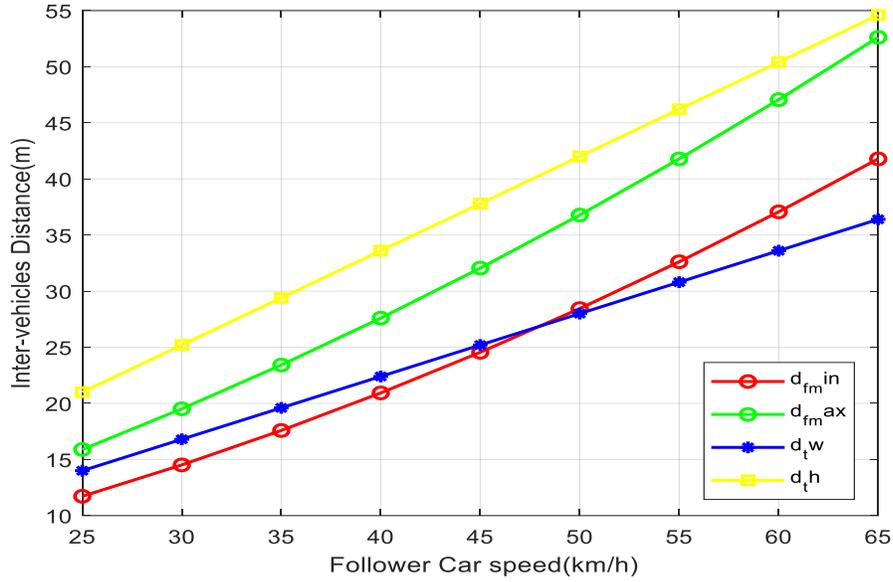

*Figure 3. Safe distance when HV speed is greater than AV at sudden braking of leading vehicle*

From the results shown in table 3 and on figure 3, the two seconds (d_tw) rule is safe since the safe distance is in the range of the safe distance of the model for the speeds up to 45 Km/hrs and dependent on how the driver is reacting to the situations. The model safe distance is higher than the two seconds rule safe distance for speeds above 45 km/hrs. If a follower driver follows at three seconds (d_th) rule, it is safe for all speed limits stated in the scenario that is higher than the model safe distance. As it is been noted in the simulation of the model that if there are higher relative differences, there is also bigger safe following distance, which allows time for the follower driver to react to any imminent incidents.

*Table 3: Safe distance Simulation data when HV speed is greater than AV at sudden braking of leading vehicle*

| Safe distance | Following car Speed(Km/hr) | | | | | | | | |
|---|---|---|---|---|---|---|---|---|---|
| (m) | 25 | 30 | 35 | 40 | 45 | 50 | 55 | 60 | 65 |
| d_f_min | 11.72 | 14.52 | 17.58 | 20.93 | 24.55 | 28.45 | 32.62 | 37.06 | 41.79 |
| d_f_max | 15.89 | 19.52 | 23.42 | 27.60 | 32.05 | 36.78 | 41.78 | 47.06 | 52.62 |

However, this scenario model indicates that depending on the spacing between the leader and the follower, the driver of the follower vehicle has to adapt to the speed of the leader if s/he can observe the differences in speed to maintain safe distance rather than obeying the rule of two seconds following distances. Therefore, the human-driven vehicle drivers have to do more than just obey the rule to care for the safety of others on the road and use defensive driving techniques for maintaining a safe following distance when driving behind driverless cars at higher speeds to avoid rear-end collision and tailgating.

4.2.2. Speed of the follower vehicle is equal to the leader vehicle's speed ($V_H = V_A$)

At each of the vehicle's speed, even though there is a fluctuation in speed for human-driven vehicles due to throttling, we supposed its speed is equal to the lead vehicle at each time they moved. The simulation data obtained using equation (16) is shown in figure 4 and table 4.

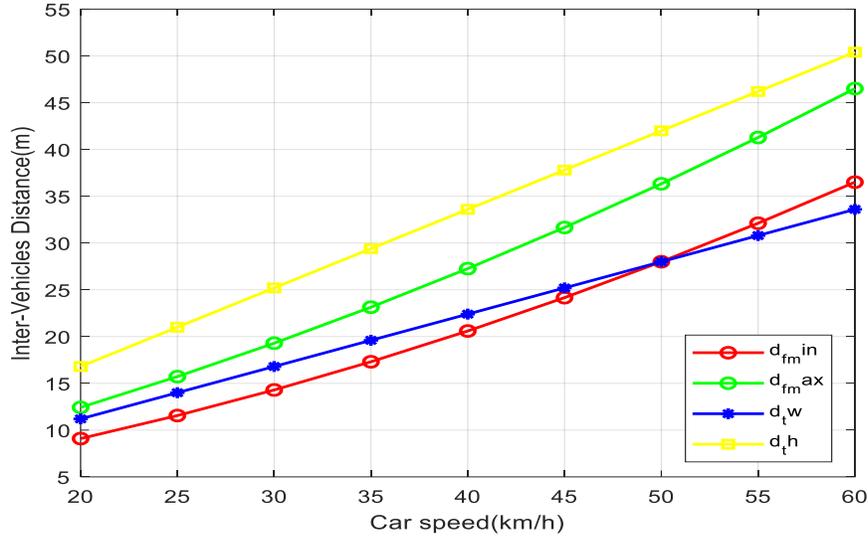

*Figure 4. Safe distance when speeds of HV and AV are equal and leading vehicle is at a sudden braking*

As shown in Table 4 about table 1 for comparison and figure 4, at lower speed the distance between vehicles is lower and gets higher with increasing speeds. The two seconds rule safe following distance is within the range of this model for human-driven vehicle and driverless interaction on the existing road at a speed limit driving condition for speeds below 50 km/hr, which has greatly been affected by how the driver is reacting to the situations. Therefore, even it is a simple truth that the faster the follower vehicle is driven, the longer it takes to stop a vehicle, the scenario simulation confirmed with this simulation data for a follower vehicle which follows a driverless car with the two-second rules of safe following distance at a speed of 50 km/hrs. If the human-driven vehicle is driven at above 50 km/hr with two seconds rule of safe following distance, the distance is lesser than the model and so it will cause rear-end collision since the time needed to stop a human-driven car is not sufficient.

*Table 4: Safe Distance Simulation Data when VH=VA at a Sudden Braking of Leading Vehicle*

| **Safe distance** | **Following car Speed(Km/hr)** | | | | | | | | |
|---|---|---|---|---|---|---|---|---|---|
| **(m)** | *20* | *25* | *30* | *35* | *40* | *45* | *50* | *55* | *60* |
| d_f_min | 9.09 | 11.56 | 14.29 | 17.31 | 20.60 | 24.16 | 28.00 | 32.12 | 36.51 |
| d_f_max | 12.43 | 15.72 | 19.29 | 23.14 | 27.26 | 31.66 | 36.33 | 41.28 | 46.51 |

From the simulation data based on this model, the existing two seconds rule of following distance works only for lower vehicle speed, less than 50 km/hr, however, the three seconds rule of safe following distance can work for all speed limits taken on the study of the safe distance.

### 4.2.3. Speed of the follower vehicle is less than the leader vehicle's speed ($V_H < V_A$)

This scenario is not the worst as to which the driver of a follower vehicle keeps sufficient spacing with the lead vehicle and is not accelerating a car. However, the driver of a follower car can accelerate to the speed equal to lead vehicle and try to adjust the speed if constrained by the lead vehicle. The simulation data is obtained using equation (21) by setting the leading vehicle speed at 60 km/hrs and invariable.

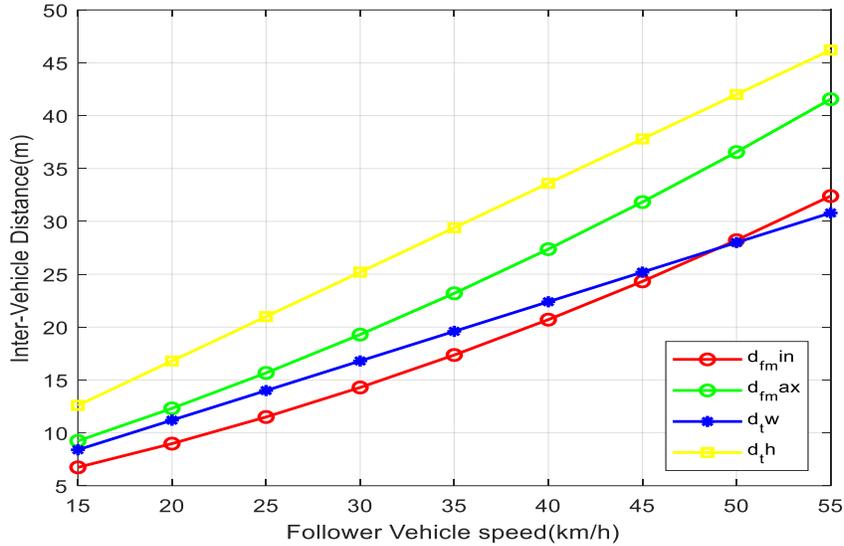

*Figure 5. Simulation data of safe distance when speeds of HV are less than AV*

As shown in figure 5, as the speed of the human-driven car speed is approaching to the speed of the driverless car, the safe distance also increased to let the driver has time to react for incidents.

*Table 5: Simulation data of safe distance when VH < VA*

| Safe distance | Following car Speed(Km/hr) | | | | | | | | |
|---|---|---|---|---|---|---|---|---|---|
| (m) | *15* | *20* | *25* | *30* | *35* | *40* | *45* | *50* | *55* |
| d_f_min | 6.74 | 8.98 | 11.50 | 14.29 | 17.36 | 20.71 | 24.33 | 28.22 | 32.39 |
| d_f_max | 9.24 | 12.32 | 15.67 | 19.29 | 23.20 | 27.37 | 31.83 | 36.56 | 41.56 |

Referring to table 1 and 5 for analysis and comparison, the three seconds rule of safe following distances are safe for all speeds limits, 20 up to 60 Km/hrs. However, the road efficiency will significantly reduce, but it is not the concern of the paper. The two seconds rules safe following distance for following a driverless car at speed greater than 45 km/hrs is less than the model. Hence, the distance to be covered with those speeds are less than the minimum safe distance of a model and causing tailgating, which leads for rear-end collision to happen. Therefore, in this scenario, the human-driven car follows a driverless car with 20 up to 45 km/hrs speeds using second rule safe following distance, however, dependent on how the driver is reacting to an imminent incident.

## 5. Summary and Conclusion

Even if there is a clear benefit in terms of reaction time and reducing accidents in front of driverless cars with the help of AEB, the results showed there is a possibility of rear-end collision to happen. Based on the four scenarios analysis of the interaction of human-driven and driverless cars using a safe distance model on a single lane dry road, it can be concluded that a fully autonomous braking system significantly affects the safe distance. Due to the variation in the braking system, the safe distance cannot be maintained, and the alerted follower driver should not be claimed and responsible for the crash as s/he follows the rules of safe following distances.

Therefore, rear-end collision may not be reduced if driverless cars are hitting the road to share with the Human-driven cars and existing traffic rules with drivers of different driving behavior. This idea is also supported by the results from accident analyses, which showed that in many cases, rear-end crashes occur in situations that are usually easy to handle (e.g. straight roads, low traffic density) [25]. In these situations, for not alerted drivers, they may not anticipate that the driver in front will brake [26] and they are thus following too closely to be able to react in time when the front vehicle suddenly brakes or stops and hence still there rear-end collision to happen which will make severe the accident.

Regarding all scenarios studied based on the new safe distance model, in following driverless cars, three seconds rules, and the new model can be used on all speed limits. However, the two seconds rule can be applied on speed limits up to 45 km/hrs, which indicates that the requirement of revising the existing car-following rules and regulations to meet the objectives of the intended purpose of driverless cars to hit the existing road.

# Reference

[1] WHO, "Global Status Report on Road," *World Heal. Organ.*, p. 20, 2018, [Online]. Available: https://www.who.int/violence_injury_prevention/road_safety_status/2018/en/.
[2] J. D. Lee, D. V. McGehee, T. L. Brown, and M. L. Reyes, "Collision warning timing, driver distraction, and driver response to imminent rear-end collisions in a high-fidelity driving simulator," *Hum. Factors*, vol. 44, no. 2, pp. 314–334, 2002, doi: 10.1518/0018720024497844.
[3] N. A. Stanton and M. S. Young, "Vehicle automation and driving performance," *Ergonomics*, vol. 41, no. 7, pp. 1014–1028, 1998, doi: 10.1080/001401398186568.
[4] S. Singh, "Critical reasons for crashes investigated in the National Motor Vehicle Crash Causation Survey.," Washington, DC.
[5] P. . Steven Goodridge, "Autonomous Driving and Collision Avoidance Technology: implications for bicyclists and pedestrian." https://www.bikewalknc.org/2018/02/autonomous-driving-and-collision-avoidance-technology/. (accessed Dec. 15, 2019).
[6] Dewesoft, "Advanced Driver Assistance Systems (ADAS) Testing." https://dewesoft.com/applications/vehicle-testing/adas (accessed Dec. 15, 2019).
[7] P. G. Gipps, "A behavioural car-following model for computer simulation," *Transp. Res. Part B Methodol.*, vol. 15, no. 2, pp. 105–111, 1981.
[8] M. Brackstone and M. McDonald, "Car-following: A historical review," *Transp. Res. Part F Traffic Psychol. Behav.*, vol. 2, no. 4, pp. 181–196, 1999, doi: 10.1016/S1369-8478(00)00005-X.
[9] A. Touran, M. A. Brackstone, and M. Mcdonald, "A collision model for safety evaluation of autonomous intelligent cruise control," vol. 31, pp. 567–578, 1999.
[10] and D. H. Treiber, M., A. Hennecke, *Microscopic simulation of congested traffic, in Traffic and granular flow'99.* Springer, 2000.
[11] K. Aghabayk, M. Sarvi, N. Forouzideh, and W. Young, "New Car-Following Model Considering Impacts of Multiple Lead Vehicle Types," doi: 10.3141/2390-14.
[12] M. Brackstone, B. Waterson, and M. Mcdonald, "Determinants of following headway in congested traffic," *Transp. Res. Part F Psychol. Behav.*, vol. 12, no. 2, pp. 131–142, 2009, doi: 10.1016/j.trf.2008.09.003.
[13] A. Gowri, K. Venkatesan, and R. Sivanandan, "Advances in Engineering Software Object-oriented methodology for intersection simulation model under heterogeneous traffic conditions," *Adv. Eng. Softw.*, vol. 40, no. 10, pp. 1000–1010, 2009, doi: 10.1016/j.advengsoft.2009.03.015.
[14] K. Venkatesan, A. Gowri, R. Sivanandan, T. E. Division, T. E. Division, and T. E. Division, "Development of Microscopic Simulation Model for Heterogenous Traffic Using Object Oriented Approach," *Transportmetrica*, vol. 4, no. 3, pp. 227–247, 2008, doi: 10.1080/18128600808685689.
[15] B. and M. S. Schoettle, "A Survey of Public Opinion About Autonomous and Self-Driving Vehicles in the U . S ., the U . K ., and Australia," Michigan, 2014.
[16] NHTSA, "Automated Vehicles for Safety," 2013. .
[17] "% F," pp. 72–78, 2012.
[18] G. Johansson and K. Rumar, "Drivers' Brake Reaction Times," *Hum. Factors J. Hum. Factors*

*Ergon. Soc.*, vol. 13, no. 1, pp. 23–27, 1971, doi: 10.1177/001872087101300104.

[19] Y. Chen and C. Wang, "Vehicle Safety Distance Warning System : A Novel Algorithm for Vehicle Safety Distance Calculating Between Moving Cars," in *IEEE 65th Vehicular Technology Conference - VTC2007-Spring*, 2007, pp. 2570–2574, doi: https://doi.org/10.1109/VETECS.2007.529.

[20] P. S. Bokare and A. K. Maurya, "Acceleration-Deceleration," in *Transportation Research Procedia*, 2017, vol. 25, pp. 4733–4749, doi: 10.1016/j.trpro.2017.05.486.

[21] A. K. Maurya and P. S. Bokare, "STUDY OF DECELERATION BEHAVIOUR OF DIFFERENT VEHICLE," vol. 2, no. 3, pp. 253–270, 2012.

[22] C. R. Greco, "Real-Time Forward Urban Environment Perception for an Autonomous Ground Vehicle Using Computer Vision and Lidar," *Engineering*, no. April, 2008.

[23] U. Environment, "Empirical Evaluation of an Autonomous Vehicle in an," *J. Aerosp. Comput. INFORMATION, Commun.*, vol. 4, no. December, 2007, doi: 10.2514/1.32839.

[24] A. Broggi *et al.*, "Extensive tests of autonomous driving technologies," *IEEE Trans. Intell. Transp. Syst.*, vol. 14, no. 3, pp. 1403–1415, 2013, doi: 10.1109/TITS.2013.2262331.

[25] D. Qu, X. Chen, W. Yang, and X. Bian, "Modeling of car-following required safe distance based on molecular dynamics," *Math. Probl. Eng.*, vol. 2014, 2014, doi: 10.1155/2014/604023.

[26] C. Li, X. Jiang, W. Wang, Q. Cheng, and Y. Shen, "A Simplified Car-following Model Based on the Artificial Potential Field," *Procedia Eng.*, vol. 137, pp. 13–20, 2016, doi: 10.1016/j.proeng.2016.01.229.